# Merging and Shifting of Images with Prominence Coefficient for Predictive Analysis using Combined Image

T.R. Gopalakrishnan Nair
Member, Advanced Imaging and Computer Vision Group
Aramco Endowed Chair, PMU, KSA
VP, RIIC, Dayananda Sagar Institutions
Bangalore, India
trgnair@gmail.com

Richa Sharma
Member, Advanced Imaging and Computer Vision Group Research and Industry Centre (RIIC), Dayananda Sagar Institutions
Bangalore, India
richarichas@yahoo.com

*Abstract*—Shifting of objects in an image and merging many images after appropriate shifting is being used in several engineering and scientific applications which require complex perception development. A method has been presented here which could be used in precision engineering and biological applications where more precise prediction is required of a combined phenomenon with varying prominence of each phenomenon. Accurate merging of intended pixels can be achieved in high quality using frequency domain techniques even though initial properties of the original pixels are lost in this process. This paper introduces a technique to shift and merge various images with varying prominence of each image. A coefficient named prominence coefficient has been introduced which is capable of making some of the images transparent and highlighting the rest as per requirement of merging process which can be used as a simple but effective technique for overlapped view of a set of images.

*Index Terms*—frequency merging, image overlapping, image shifting, prominence coefficient.

## I. Introduction

Fundamentally, Images are a collection of pixels and each pixel is represented by some values depending on the type of the image, dealing them at intensity level is insufficient for precision management. In gray level digital images, pixels contain intensity information and we perceive and analyse an image based on the changes in shades of color intensities or frequencies [1]. An object in an image is identified by its color, shape and texture. In the field of image processing, color information can be processed easily as the image is stored as a collection of pixels of various colors or intensities. Intensity slicing is a very simple and effective technique which at times can give extremely unbelievable and surprisingly desired results in no time

Each object in an image generates a unique spectrum of frequencies in frequency domain. An image with many such objects contains a collection of all those frequencies. This paper demonstrates the feasibility of creating accurate high-fidelity images through merging and shifting, for prediction purposes in biology and solid mechanics. In order to cater to the needs of complete information of image, it is necessary and better to merge the frequencies along with phase to get overlapped view of the objects or images. Such methods are required for several practices in astronomy, radiology, body implant predictions, fracture mechanics prediction etc. [13].

In the field of image processing, color/ gray level intensity information are easy to process because the images are stored as a collection of pixels of various colors or intensities. A point in an image is called an edge point if it shows sudden change in intensity and we can visualize and identify various objects in an image because of their color, shapes and texture. It is the rate of change of intensity values which gives the illusion of an object.

In literature, Image merging has been used for many reasons like resolution improvement or for segmentation purposes. A novel image restoration algorithm to deblur the image without estimating the image blur by merging differently blurred multiple images in the spectrum domain using the fuzzy projection onto convex sets (POCS) can be found in [2]. Statistical Region Merging (SRM) and the Minimum Heterogeneity Rule (MHR) have also been used for object merging [3]. The SRM segmentation method not only considers spectral, shape, and scale information, but also has the ability to cope with significant noise corruption and handle occlusions. The MHR used for merging objects takes advantage of its spectral, shape, scale information, and the local and global information. A high level of fusion quality in global information can be achieved by a novel self-adaptive weighted average fusion scheme based on standard deviation of measurements to merge IR and visible images [4].

Merging of different data sets in digital image processing is mainly used to improve the visual and analytical quality of the data [9]. The data may be of different types such as satellite imagery from the same sensor but with different resolution, satellite imagery from different sensors with varying resolution, digitized aerial photography and satellite imagery or satellite imagery with ancillary information can be merged. There are many techniques for merging like Principal Component, IHS, and Brovey Transform. A technique for multi-image fusion in one-pass through overlapping input images which restores and reconstructs the scene radiance field can be found in [10]. The technique is effective because it maximizes fidelity based on a comprehensive end-to-end system model that accounts for scene statistics, acquisition blurring, sampling, and noise.

Multisensor image fusion is the process of combining relevant information from two or more images into a single image. The resulting image will be more informative than any of the input images. Image fusion has become a common term used within medical diagnostics and treatment too. The term is used when certain portions of multiple images of a patient are registered and overlaid or merged to provide additional combined information. Fused images may be created from multiple images from the same imaging modality [5], or by combining information from multiple modalities [6], such as magnetic resonance image (MRI), computed tomography (CT), positron emission tomography (PET), and single photon emission computed tomography (SPECT). These images serve different purposes in radiology and radiation oncology. For example, CT images are used more often to ascertain differences in tissue density while MRI images are typically used to diagnose brain tumors. Merging prostate imaging has also been used to identify the location and aggressiveness of prostate cancer [7].

The increasing availability of space borne sensors gives a motivation for different image fusion algorithms in radiology and radiation oncology. Several situations in image processing require high spatial and high spectral resolution in a single image. Most of the available equipments are not capable of providing such data convincingly [12]. The image fusion techniques allow the integration of different information sources. The fused image can have complementary spatial and spectral resolution characteristics. However, the standard image fusion techniques can distort the spectral information of the multispectral data while merging. Data fusion method for land cover classification that combines remote sensing data at a fine and a coarse spatial resolution can be found in [8]. This classifier uses all image information (bands) available at both fine and coarse spatial resolutions by stacking the individual image bands into a multidimensional vector.

Traditionally, image merging is used for enhancing resolution of an image or for 3D segmentation. This paper is not about that but it is about merging two or more images to get the collective view of the image. The authors have earlier presented the concept of object or images merge in frequency domain [14]. This paper is one step further in that direction as the study of effects of prominence coefficient and appropriate shifting in frequency domain before merging has been investigated here. Merging of images discussed in this paper can be used in many applications, for example, solid property predictions, creating currency, body implant, compression fractures, fast viewing or creative editing. In biological systems multi cellular growth patterns created separately can be viewed as single slide for studies of growth which is required for damage characteristics prediction using the technique presented here.

## II. Spectral treatment

The repetitive nature or the frequency characteristics of images can be analyzed using spectral decomposition methods like Fourier analysis. In an RXC (*Row × Column*) digital image, positions *u* and *v* indicate the number of repetitions of the sinusoid in those directions. Therefore the wavelengths along the column and row axes are

$\lambda_u = \frac{C}{u}$ and $\lambda_v = \frac{R}{v}$ pixels,

and the wavelength in the wavefront direction is

$\lambda_{wf} = \sqrt{\left(\frac{C}{u}\right)^2 + \left(\frac{R}{v}\right)^2}$.

The frequency is the fraction of the sinusoid traversed over one pixel,

$\omega_u = \frac{u}{C}$, $\omega_v = \frac{v}{R}$, and

$\omega_{wf} = 1 \bigg/ \sqrt{\left(\frac{C}{u}\right)^2 + \left(\frac{R}{v}\right)^2}$ cycles.

The wavefront direction is given by

$$\theta_{wf} = \tan^{-1}\left(\frac{\omega_v}{\omega_u}\right) = \tan^{-1}\left(\frac{vC}{uR}\right).$$

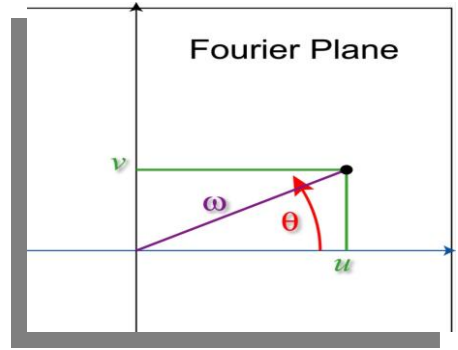

Fig. 1. Details of Fourier Plane

## III. IMPLEMENTATION

This section describes the algorithms used for precision merging and shifting. We have created a prototype implementation for object shifting and merging using the following algorithms:

***Shifting and Merging in spatial domain:***

Suppose there are '*n*' images to be merged each of size RXC (*Row × Column*). Shifting of image '*i*' can be represented by the following formulae:

$i_{shifted}(x,y) = i(x+S, y+S);$ where $0 < x \leq R-S$ (1)
$0 < y \leq C-S$

Otherwise

$i_{shifted}(x,y) = i(R-x, C-y);$ where $R-s+1 < x < R$
$C-s+1 < y < C$

Where '*S*' is the displacement in terms of number of pixels.

A. Algorithm

Initialization:

Set the value for number of images/layers 'n'
Align the images or layers so that number of pixels is same.
Normalize the image values i(x,y) to 0-1
Identify the images which require a shifting of objects. Shift objects using equation (1).
**While** all pixels(x,y) in the image are not seen **do**
ADD corresponding intensities of all n images
$$Result(x, y) = i_1(x, y) + \ldots + i_n(x, y)$$

***Shifting and Merging in frequency domain:***

In frequency spectrum of an RXC digital image, positions $u$ and $v$ indicate the number of repetitions of the sinusoid in those directions. We scale each sinusoid with a prominence coefficient(PC) before integration for perception control. As per definition of Fourier Transform, frequency spectrum of image '$i$' can be represented by the following formulae:

$$FFT(i(x, y)) = I(u,v) = \sum_x \sum_y i(x, y)\exp[-j2\pi(ux/R + vy/C)] \quad (2)$$

Hence,

$$I(u,v) = \sum_x \sum_y i(x, y)[\cos(\theta) - j\sin(\theta)] \quad (3)$$

Where,

$$\theta = 2\pi(ux/R + vy/C) \quad (4)$$

Same can be rewritten as:

$$I(u,v) = \sum_x \sum_y i(x, y)[R_\theta - I_\theta] \quad (5)$$

Where, $R$ and $I$ stand for real part and imaginary part of Fourier Spectrum.

Suppose there are n images to be merged and each image pixel p can be represented as $p_{xyn}$ in 3D domain of n images and its corresponding frequency $P_{uvn}$ can be calculated by the following equation:

$$P_{uvn} = a_n \sum_x \sum_y i_n(x, y)[R_\theta - I_\theta] \quad (6)$$

And

$$P_{integral} = \sum_n P_{uvn} \quad (7)$$

Where $a_1, a_2 \ldots a_n$ are prominence coefficients.
Therefore,

$$P_{integral} = a_1 \sum_x \sum_y i_{1(x,y)}[R_\theta - I_\theta] + \ldots$$
$$\ldots + a_n \sum_x \sum_y i_{n(x,y)}[R_\theta - I_\theta] \quad (8)$$

Actual values of these prominence coefficients will be adjusted by visual perception depending on the application and requirement. The effect of prominence coefficient has been discussed in section IV of this paper.

It may also require shifting the objects in the images before merging in order to get a meaningful and useful resultant image. The translation property of the Fourier trasform has been used to shift the objects. Shifting in frequency domain can be achieved by using the following formula:

$$i_{shifted} = IFFT(FFT(i)\exp(-j2\pi wS)) \quad (9)$$

Where,
'$w$' is the vector of the frequencies used in the exponential that allow the shift:

*w=[0:floor((numel(I)-1)/2)-ceil((numel(I)-1)/2):-1]/ numel(I)*

'$S$' is the displacement in terms of number of pixels.

B. Algorithm

**Initialization:** Set the value for number of images/layers 'n'. Align the images or layers so that number of samples is same. Normalize the image values i(x,y) to 0-1.

Identify the images which require a shifting of objects. Shift objects using equation (9).

**While** all pixels(x,y) in the images are not seen **do**
ADD corresponding density of frequencies of all n images
using following formula:

$$Result(u,v) = a_1 FFT(i_1(x,y)) + ...... + a_n FFT(i_n(x,y))$$

Provide higher prominence coefficient values to the images we wish to be more prominent in merged domain.
Take inverse FFT to visualize the merged image.

## IV. RESULT AND ANALYSIS

The experiments were aimed at developing a system to merge images in such a way that all images can be seen together in high fidelity without much loss of information because of the overriding intensity factors.

The comparative study was performed over ten images for shift and merge in frequency domain by using Algorithm A and B discussed in section III. It is found by visual inspection that merging in spatial domain as well as frequency domain both gives similar results if the objects in input images are spatially separate or non overlapping from each other (Fig.2) whereas merging in frequency domain out performs if objects in the images are getting overlapped in merged image. We have not gone for quantitative methods to further check for error occurred while imaging, as visual inspection itself gave satisfactory evidence. Merging in frequency domain is capable of keeping fine details of all input images (Fig.3). This technique can be useful in surveillance where the observer is supposed to see multiple pictures coming from many cameras at a time as it gives ease to the observer to view multiple images at a time in a single screen. If we merge more than 2 images, we can get further compressed view.
Fig. 4 shows the effects of prominence coefficient on X- ray image of fractured bone. It shows that while merging we can fix prominence of each image according to the requirement.

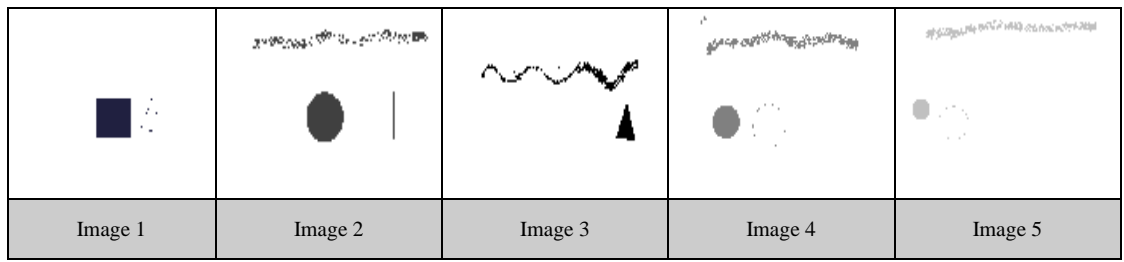

(a)

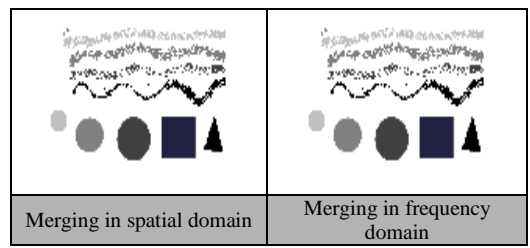

(b)
Fig. 2. (a). Artificially generated Input Images (b). Merged image generated by Algorithm A and B for n=5. Output quality looks same in both cases.

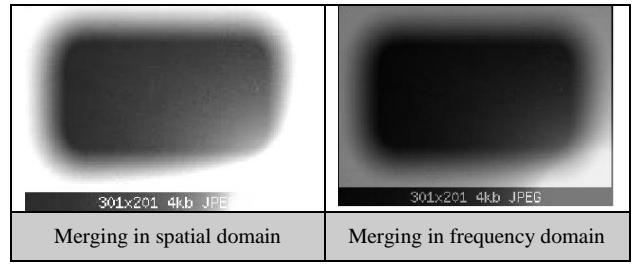

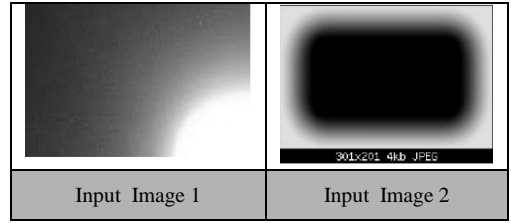

Fig. 3. (a). Images to be merged (b). Merging using Algorithm A and Algorithm B for n=2. Faded circular part of image 1 and text part of image 2, both are more clearly seen in frequency merged image.

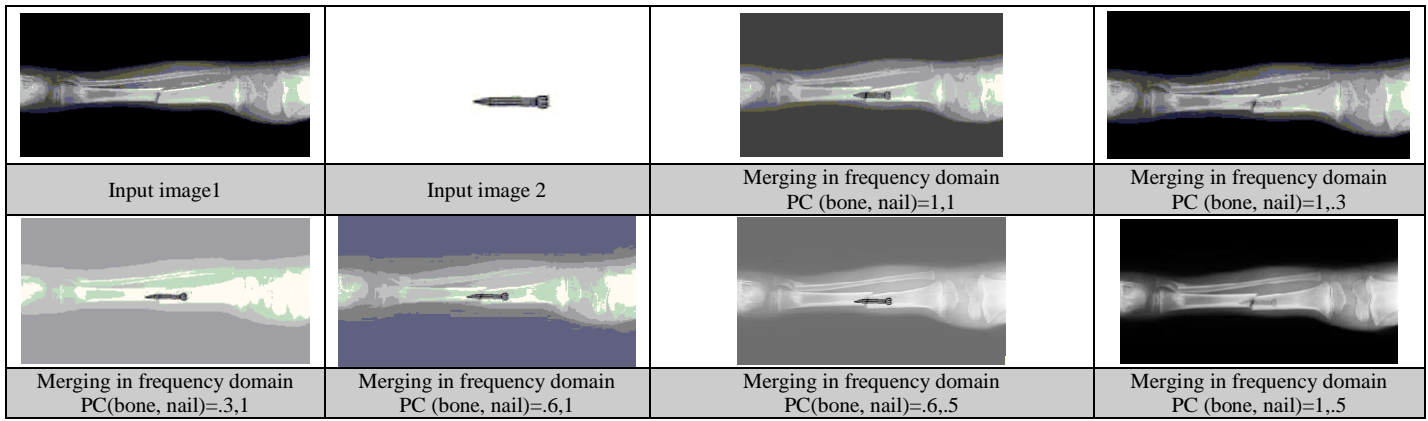

Fig. 4. Natural Images merged using Algorithm B for n=2 with various values of prominence coefficient.

## V. CONCLUSION

A simple but effective technique for object shifting and merging using linear integration of spectral density with varying prominence has been introduced in the paper. A coefficient named prominence coefficient has been discussed which is capable of making some of the images transparent and highlighting the rest of the images as per requirement of the application in merging process. This can be used to improve the quality of merged image and is simple but effective technique for producing overlapped view of a set of images.

The paper demonstrates a technique to modify prominence of each image and shift them individually before merging according to our needs in frequency domain. The images merged with varying prominence have displayed improved precision, which is capable of being used in various applications like fracture prediction in solids, multicellular integration in biology and visualization of body implants.